\documentclass[runningheads]{llncs}
\usepackage{graphicx}
\usepackage{booktabs}
\usepackage[colorinlistoftodos]{todonotes}
\usepackage[inline]{enumitem}

\usepackage{chngcntr}
\counterwithout{figure}{chapter}
\counterwithout{table}{chapter}

\begin{document}
\title{Commonsense Knowledge in Wikidata}
%
%
\author{Filip Ilievski\inst{1} \and
Pedro Szekely\inst{1} \and
Daniel Schwabe\inst{2}}
\authorrunning{Ilievski, Szekely, and Schwabe}
%
\institute{Information Sciences Institute, University of Southern California \email{\{ilievski,pszekely\}@isi.edu} \and Dept. of Informatics, Pontificia Universidade Cat\'olica Rio de Janeiro \\ \email{dschwabe@inf.puc-rio.br}
}
\maketitle              
\setcounter{footnote}{0}
\setcounter{table}{0}

\begin{abstract}
Wikidata and Wikipedia have been proven useful for reasoning in natural language applications, like question answering or entity linking. Yet, no existing work has studied the potential of Wikidata for commonsense reasoning. This paper investigates whether Wikidata contains commonsense knowledge which is complementary to existing commonsense sources. Starting from a definition of common sense, we devise three guiding principles, and apply them to generate a commonsense subgraph of Wikidata (\emph{Wikidata-CS}). Within our approach, we map the relations of Wikidata to ConceptNet, which we also leverage to integrate Wikidata-CS into an existing consolidated commonsense graph. Our experiments reveal that: 1) albeit Wikidata-CS represents a small portion of Wikidata, it is an indicator that Wikidata contains relevant commonsense knowledge, which can be mapped to 15 ConceptNet relations; 2) the overlap between Wikidata-CS and other commonsense sources is low, motivating the value of knowledge integration; 3) Wikidata-CS has been evolving over time at a slightly slower rate compared to the overall Wikidata, indicating a possible lack of focus on commonsense knowledge. Based on these findings, we propose three recommended actions to improve the coverage and quality of Wikidata-CS further.

\end{abstract}
\section{Introduction}

Common sense is ``the basic ability to perceive, understand, and judge things that are shared by nearly all people and can be reasonably expected of nearly all people without need for debate''~\cite{gunning2018machine}. For instance, humans typically know that the political opposition is an opposite of the government, that hunger causes one to eat, and that 
if one walks in the rain one gets wet.
Possessing such commonsense knowledge is important for both humans and machines in order to fill gaps in communication, and fulfill tasks such as entity recognition and linking from text, question answering, and planning.
Yet, understanding common sense is difficult for machines. Even with the recent progress of language models such as BERT~\cite{devlin2018bert} and GPT-2~\cite{radford2019language}, which have been able to perform very well on a number of tasks with enough training,\footnote{For example: \url{https://leaderboard.allenai.org/socialiqa/submissions/public}} the correct answer is often given for wrong reasons~\cite{ettinger2020bert}. The utterances produced are syntactically sound, but may lack plausibility. For instance, GPT-2 complements the following prompt \textit{`if you break a bottle that contains liquids, some of the liquid will (other things being equal) probably...'} with \textit{`...wind up 300 meters away'}~\cite{marcus2020next}.

Commonsense graphs like ConceptNet~\cite{speer2017conceptnet} and ATOMIC~\cite{sap2019atomic} provide relevant knowledge that can be used to enhance the ability of language models to reason on downstream tasks. Unfortunately, these are largely incomplete, e.g., while ConceptNet contains information that a barbecue can be located in a garage, it is unable to infer that they are also common in other outdoor places, like parks, nor it has information on the expectations from such an event.

According to \cite{storks2019commonsense}, common knowledge graphs (KGs) derived from Wikipedia, such as Wikidata~\cite{vrandevcic2014wikidata} or YAGO4~\cite{tanon2020yago}, provide knowledge which is `often required to achieve a
deep understanding of both the low- and high-level concepts found in language'. In addition, Wikipedia has been used by a large number of systems for downstream reasoning tasks~\cite{hovy2013collaboratively}. As the largest and highest-quality structured counterpart of Wikipedia~\cite{farber2015comparative}, Wikidata holds the promise of containing useful commonsense knowledge. Yet, to our knowledge, no existing work has studied the commonsense coverage of Wikidata.

In this paper, we investigate whether Wikidata contains commonsense knowledge and whether that is complementary to existing commonsense knowledge graphs. The contributions of the paper as as follows:
\begin{enumerate*}
    \item we formulate three key principles for distinguishing commonsense knowledge from the remaining one in Wikidata, starting from three key properties of commonsense knowledge and from a survey of existing commonsense knowledge graphs (Section \ref{ssec:principles}). These principles dictate that commonsense knowledge concerns well-known concepts and general-domain relations.
    \item Based on these principles, we design and implement computational steps to extract a commonsense subgraph from Wikidata which we refer to as \textit{Wikidata-CS} in the remainder of this paper (Section~\ref{ssec:approach}). During these steps, we also map relations in Wikidata to relations in ConceptNet.
    \item We leverage this mapping to integrate Wikidata-CS into the Commonsense Knowledge Graph (CSKG)~\cite{ilievski2020consolidating}, which already contains well-known commonsense sources, such as ATOMIC and ConceptNet (Section \ref{ssec:cskg}).
    \item We perform quantitative and qualitative analysis of the resulting subgraph (Section~\ref{ssec:stats}). In addition, we compute overlaps between Wikidata-CS and other resources included in CSKG, including ConceptNet and WordNet (Section~\ref{ssec:cskg_graphs}).
    \item We perform the same experiments with three different versions of Wikidata from 2017, 2018, and 2020, and compare the results (Section \ref{ssec:time}). This allows us to quantify the evolution of commonsense knowledge in Wikidata over time.
    \item  In Section ~\ref{ssec:discussion}, we reflect on the findings from our experiments and propose recommended actions for further inclusion of commonsense knowledge in Wikidata.
\end{enumerate*}

\section{Related Work}
\label{sec:rel_work}


We review: 
\begin{enumerate*}
\item well-known commonsense KGs
\item prior works on reasoning with Wikidata or Wikipedia over text
\item studies of completeness of Wikidata
\end{enumerate*}.

\textbf{Commonsense KGs} such as ConceptNet 
and ATOMIC 
are popular and have been utilized by downstream reasoners~\cite{ma2019towards}. Lexical resources, like WordNet~\cite{miller1995wordnet} and FrameNet~\cite{baker1998berkeley}, capture commonsense knowledge about concepts and frames, respectively. Moreover, sources like Visual Genome~\cite{krishna2017visual} which have been originally proposed for a different purpose (image captioning and visual recognition), have recently been recognized as sources of commonsense knowledge. Commonsense knowledge can also be extracted from documents~\cite{tandon2017webchild,mishra2017domain}, query logs~\cite{romero2019commonsense}, or quantities~\cite{elazar2019large}. A recent resource, called the Commonsense Knowledge Graph~\cite{ilievski2020consolidating}, consolidates many of these resources into a single KG. The complementarity of these sources motivates their integration, but also reveals that they are still largely incomplete. Wikidata, as one of the richest public KGs, holds a promise to enrich the set of recorded commonsense facts even further. 

A recent idea is to use language models, like BERT~\cite{devlin2018bert} and GPT-2~\cite{radford2019language}, as knowledge bases, due to their inherent ability to produce a fact for any input prompt. Still, they often exhibit shallow understanding of the world~\cite{ettinger2020bert}. Integration with KGs like Wikidata or ConceptNet may increase their robustness~\cite{ma2019towards}.

\textbf{Reasoning with Wikipedia and Wikidata} Wikipedia and Wikidata serve as sources of background knowledge in natural language processing tasks~\cite{hovy2013collaboratively}, e.g., as a repository of entities to link to, or as a source of contextual information to help linking entities in text~\cite{mulang2019context,cetoli2019neural,delpeuch2019opentapioca}. The work by Suh et al. \cite{suh2006extracting} attempts to extract commonsense knowledge from Wikipedia.
As far as we are aware, there is no comprehensive proposal to extract commonsense knowledge from Wikipedia or Wikidata, or to study their strengths and weaknesses for this purpose.

\textbf{Studies of completeness of Wikidata} Several papers study the completeness of Wikidata~\cite{darari2017cool,balaraman2018recoin,luggen2019non}. Luggen et al.~\cite{luggen2019non} provide an approach to estimate class completeness in knowledge graphs, and use Wikidata as a use case. They note that some classes in Wikidata, like Painting, are more complete than others, such as Mountain. In addition, they also quantify the evolution of Wikidata over time. Similarly, we also study the completeness of Wikidata and its richness over time, albeit focusing on its coverage of commonsense knowledge. 

\section{Extraction of Commonsense Knowledge from Wikidata}

\subsection{Principles of commonsense knowledge} 
\label{ssec:principles}

Common sense is ``the basic ability to perceive, understand, and judge things that are shared by nearly all people and can be reasonably expected of nearly all people without need for debate''~\cite{gunning2018machine}. From this definition, we can infer that commonsense knowledge: 1) concerns conceptual rather than instance-based information; 2) is primarily about commonly known observations; 3) targets general-domain information. We expand these three aspects into three guiding principles for our approach, which would allow us to define a commonsense subset of the knowledge in a general KG such as Wikidata.

\begin{enumerate}[label=P\arabic*]
 \item \textbf{Concepts, not entities} The primary principle of commonsense knowledge draws on the distinction between concept- and named-entity-level (instance) knowledge. Generally speaking, most concept-level knowledge is common sense, whereas most named-entity-level knowledge is not. 
 The fact that \emph{houses} have \emph{rooms} is commonsense knowledge, as it common and widely applicable;
 the fact that the \emph{Versailles Palace} has \emph{700 rooms} is not, as it concerns a particular instance and cannot be expected by most people.
 Thus, principle P1 is that commonsense knowledge has to be about concepts.

\item \textbf{Commonness} The second principle (P2) of commonsense knowledge is its `commonness': it is knowledge about well-known concepts that is shared among most human beings. The fact that a \emph{container} (Q987767) is used for \emph{storage} (Q9158768) is a common fact, whereas the fact that \emph{noma} (Q994794) is a subclass of \emph{aphthous stomatitis} (Q189956) is fairly unknown.

\item \textbf{General-domain knowledge} The third principle is that commonsense knowledge is about general-domain information rather than expert knowledge about a specific domain like chemistry or biology. Notably, even within a knowledge type, some relations  describe general information, whereas others require expert knowledge. For instance, considering meronymy relations, we observe that \emph{part of} describes well-known facts (e.g., wheel is part of a car), whereas \emph{cell component} focuses on biological knowledge (e.g., \emph{cholesterol} has component \emph{cell membrane}). As a third principle (P3), we aim to distinguish between domain-specific and general-domain knowledge.
\end{enumerate}

\subsection{Approach}
\label{ssec:approach}

Next, we apply P1-P3 to select commonsense knowledge from Wikidata.

\textbf{1. Excluding named entities (P1)} In practice, Wikidata does not make a clear distinction between concepts and named instances through its structured knowledge. The relation \textit{instance of} (\texttt{P31}) would intuitively be useful for this; yet, it often expresses an \textit{is-a} relation between concepts, similar to \textit{subclass of} (\texttt{P279}). For instance, Wikidata states that \textit{surgeon} is an instance of \textit{medical profession}, and a subclass of \textit{medical specialist}. Leveraging the \texttt{rdf:type} relation from another public ontology, such as DBpedia, is a possible direction, yet this strategy would be limited to the set of nodes that are mapped between Wikidata and DBpedia.
Hence, we follow a different route. The convention of Wikidata stipulates that the labels of named entities should be capitalized, whereas the ones for concepts should not.\footnote{\url{https://www.wikidata.org/wiki/Help:Label}} Following this rule, we employ a simple heuristic of selecting edges where both nodes have alphanumeric labels starting with a lowercase letter. We expand this rule and filter out labels that contain any capital letter, to remove entities with labels like ``graf Nikolai Aleksyeevich Sheremetev''. This procedure implicitly excludes nodes without English labels.

\begin{table}[!t]
	\centering
	\caption{Number of edges and representative examples for the top 20 relations.}
	\begin{tabular} {l c r}
		\toprule
		 \bf Relation & \bf \#edges & \bf Examples  \\
		 \midrule
		subclass of (P279) & 172,535 & saxophone - woodwind instrument  \\ 
        instance of (P31) & 141,499 & happiness - positive emotion \\
        part of (P361) & 9,118 & shower - bathroom \\
        different from (P1889) & 7,767 & vein - artery  \\
        has part (P527) &  6,252 & senses - touch \\ 
        cell component (P681) &  5,607 & cholesterol - cell membrane  \\
        property constraint (P2302) & 5,180 & votes received - integer constraint \\
        facet of (P1269) & 4,792 & wind - weather \\
        strand orientation (P2548) & 4,345 & sac-1 - forward strand \\
        use (P366) & 3,045 & crystal ball - psychic reading \\ 
        opposite of (P461) & 3,028 &  political opposition - government \\
        properties for this type (P1963) & 2,382 & human - date of birth \\
        molecular function (P680) & 2,369 &  protein kinase - kinase activity \\
        see also (P1659) & 2,344 & position held - member of \\
        sport (P641) & 2,338 & head stand - gymnastics \\
        followed by (P156) & 2,244 & middle school - secondary school \\
        follows (P155) & 2,234 & queen - jack \\
        material used (P186) & 2,047 & ice cream cone - wafer \\
        is a list of (P360) & 1,914 & list of major opera composers - human \\
        Wikidata property (P1687) & 1,746 & president - head of government \\
		\bottomrule
	\end{tabular}
	\label{tab:statsx}
\end{table}

\textbf{2. Characterizing commonness (P2)} We argued that commonsense facts concern common concepts. Wikidata-based metrics of frequency or popularity, such as PageRank, cannot be used to estimate commonness, as they inherit the bias towards topics that are heavily represented in Wikidata (e.g., entertainment or science). Instead, we approximate commonness by frequencies of word and phrase usage that have been pre-computed over an independent corpus~\cite{robyn_speer_2018_1443582}.\footnote{\url{https://pypi.org/project/wordfreq/}} Here, we assume that frequently occurring words and phrases refer to well-known concepts. 
According to this tool, the frequency of a common word, like \emph{storage}, is much higher than that of a relatively unknown word, such as \emph{noma} (3.39e-05 compared to 3.24e-07). We select edges where both the subject and the object labels have usage frequency above an empirically determined threshold of $1e-06$. 

\begin{figure}[t]
    \centering
    \includegraphics[width=\textwidth]{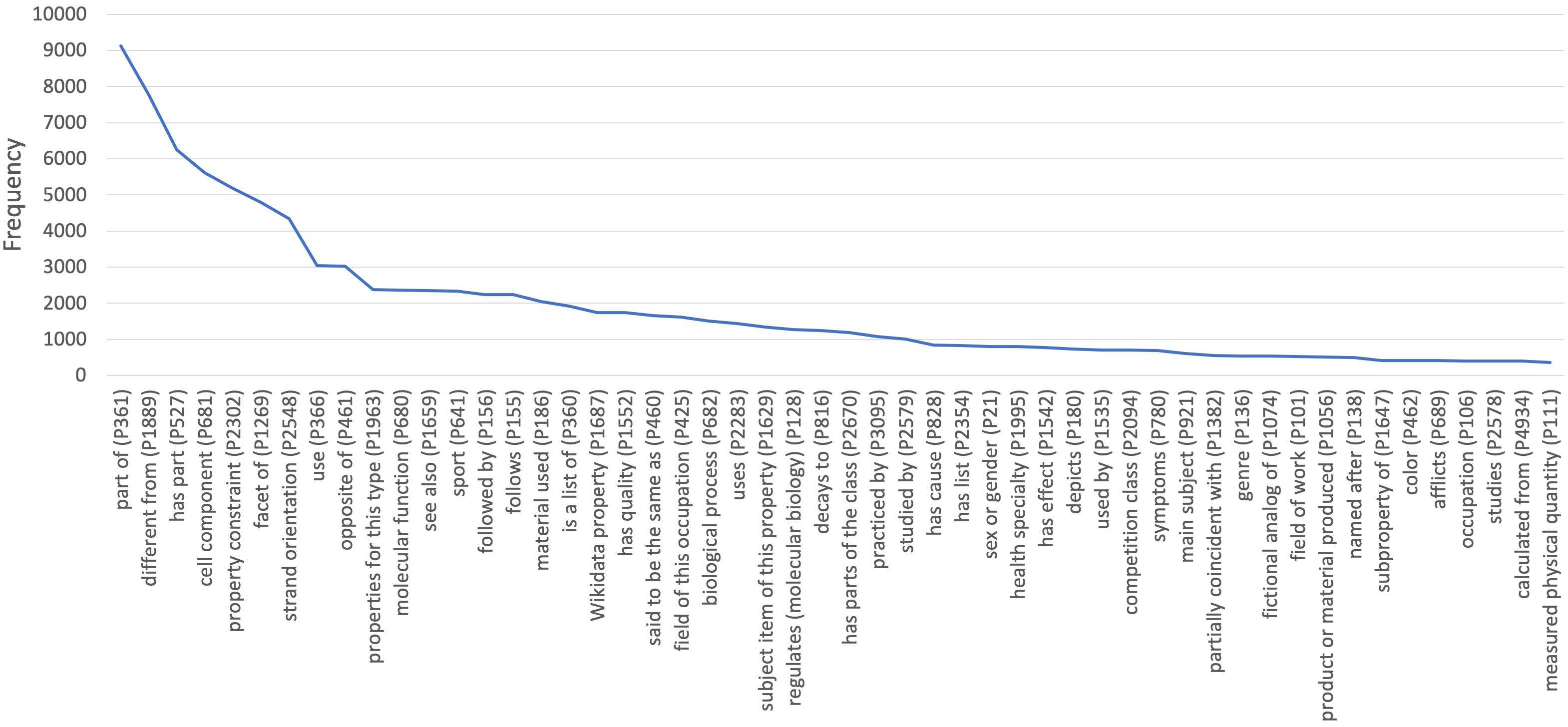}
    \caption{Frequency distribution of the 50 most frequent remaining relations. For readability, we exclude P31 and P279, as these are much more populated than the rest.}
    \label{fig:freqs}
\end{figure}

\begin{table}[!t]
	\centering
	\caption{Mapping of relations between Wikidata and ConceptNet. 
	The Wikidata relations prefixed with `*' are inverse to the relation in ConceptNet.}
	{\footnotesize
	\begin{tabular} {l c l}
		\toprule
		 \bf Category & \bf ConceptNet & \bf Wikidata  \\
		 \midrule
		 distinctness & /r/DistinctFrom & different from (P1889) \\ \hline
		 antonymy & /r/Antonym & opposite of (P461) \\ \hline
		 synonymy & /r/Synonym & said to be the same as (P460) \\ \hline
		 similarity & /r/SimilarTo & partially coincident with (P1382) \\ \hline
		 derivation & /r/DerivedFrom & named after (P138), fictional analog of (P1074) \\ \hline
		 inheritance & /r/IsA & instance of (P31), subclass of (P279), subproperty of \\ 
		 & & (P1647) \\ \hline
		  meronymy & /r/PartOf & part of (P361), *has part (P527), *has parts of the \\
		  & & class (P2670)\\ \hline
		  material & /r/MadeOf & material used (P186), is a list of (P360), *has list \\
		  & & (P2354) \\
		  \hline
		  attribution & /r/CreatedBy & *product or material produced (P1056) \\ \hline
		  utility & /r/UsedFor & use (P366), *uses (P2283), used by (P1535) \\ \hline
          properties & /r/HasProperty & color (P462), has quality (P1552), properties of this\\
          & &   type (P1963), Wikidata property (P1687), sex or\\ 
          & & gender (P21) \\ \hline
          causation & /r/Causes & *has cause (P828), has effect (P1542), symptoms \\
          & & (P780) \\ \hline
          ordering & /r/HasPrerequisite & *followed by (P156), follows (P155) \\ \hline
          context & /r/HasContext & facet of (P1269), field of this occupation (P425),\\
          & &  health specialty (P1995), main subject (P921),\\
          & & competition class (P2094), genre (P136), studied  \\
          & & by (P2579), field of work (P101), afflicts (P689), \\
          & & *practiced by (P3095), depicts (P180), sport (P641) \\
          \hline
          other & /r/RelatedTo & see also (P1659), subject item of this property \\
          & & (P1629) \\ 
		\bottomrule
	\end{tabular}
	}
	\label{tab:mapping}
\end{table}

\textbf{3. Excluding domain knowledge (P3)} The initial two steps yield 420,822 edges, involving 414 edge types and describing 194,595 nodes. Table~\ref{tab:statsx} presents the number of occurrences for the 20 most frequent edge types, together with a representative example edge for each type. 
By analyzing the frequency distribution of the remaining relations, we observe that the frequency quickly decays. The 50th most common relation describes less than 500 edges, and their frequency plot becomes relatively flat (Figure \ref{fig:freqs}). Hence, we focus on the 50 most frequent relations and distinguish the remaining knowledge by manually mapping them to relations in ConceptNet v5.7.\footnote{\url{https://github.com/commonsense/conceptnet5/wiki/Downloads}} These account for 409,775 edges, which is 97.4\% of the total set of edges available at this point.\footnote{In the future, we intend to consolidate the remaining statements of Wikidata by mapping them to ConceptNet relations as well.} 

The main guideline for this mapping was to exclude properties which are meant to describe domain-specific information, such as \textit{strand orientation} (\texttt{P2548}).
The mapping was performed independently by two authors of this paper. In all cases, the annotators agreed on whether the relation describes general-domain knowledge. In 9 cases, the annotators disagreed on which ConceptNet relation is the most appropriate to map to. Typically, this meant that ConceptNet lacks a relation with the same specificity, forcing the annotators to opt for a more generic relation, such as \texttt{/r/HasContext}. The disagreements were resolved through a joint discussion and examination of exemplar edges in Wikidata.

The resulting mappings are shown in Table~\ref{tab:mapping}. 44 out of the top 50 relations were mapped to existing relations in ConceptNet, yielding 388,250 edges. The remaining six relations are either biology domain-specific: cell component (P681), strand orientation (P2548), molecular function (P680), biological process (P682); physical domain-specific: decays to (P816); or ontological: property constraint (P2302). The mapping shows that some ConceptNet properties (e.g., \texttt{/r/Antonym}) have a single counterpart in Wikidata (\textit{opposite of}), while others (e.g., \texttt{/r/HasContext}) map to several properties, often with more specific meanings (e.g., \textit{genre}, \textit{sport}). This might reveal an opportunity to enrich the specificity of relations in ConceptNet with more detailed ones as in Wikidata.\footnote{For some ConceptNet relations, like \texttt{/r/PartOf} and \texttt{/r/HasProperty}, a similar proposal to add detail comes from WebChild~\cite{tandon2017webchild}.} Some relations in ConceptNet (e.g., \texttt{/r/MotivatedByGoal}) may have no counterpart in Wikidata, and others map to a relation which is very sparse for common concepts. For instance, \texttt{/r/AtLocation} maps to \textit{location}, which is well-populated for named entities in Wikidata, but only ranks 72nd with 159 occurrences in our commonsense subset. 
These observations reveal a knowledge gap in Wikidata. In several cases, the relation in Wikidata is inverse to that in ConceptNet, e.g., \textit{has part} to \texttt{/r/PartOf} and \textit{has cause} to \texttt{/r/Causes}. We analyze the overlap between ConceptNet and Wikidata further in Section~\ref{ssec:cskg_graphs}.

Finally, assuming that domain-specific relations involve domain-specific nodes, we construct a set of `blacklist' nodes found in these relations. We ensure that the remaining edges do not contain these domain-specific nodes. This allows us to filter out nodes like protein (\texttt{Q8054}), which has over 172 thousand incoming edges, typically from child proteins.



\subsection{Integration in the Commonsense Knowledge Graph}
\label{ssec:cskg}

The Commonsense Knowledge Graph (CSKG)~\cite{ilievski2020consolidating} is an existing resource that consolidates information from seven commonsense sources, including ConceptNet, Roget~\cite{kipfer2005roget}, Visual Genome~\cite{krishna2017visual}, WordNet~\cite{miller1995wordnet}, and Wikidata. 
It is represented using the KGTK~\cite{ilievski2020kgtk} format with 10 columns, including the core elements of an edge (\texttt{id}, \texttt{node1}, \texttt{relation}, and \texttt{node2}), their labels (e.g., \texttt{node1;label}, and provenance information about an edge (\texttt{source} and \texttt{sentence}).
Regarding Wikidata, CSKG includes all the edges involving the inheritance (\texttt{P279}) relation. 

We integrate the commonsense subset of Wikidata presented in this paper into CSKG. For this purpose, we adapt its columns to match those specified by CSKG. The columns for which we lack information, such as \texttt{sentence}, are left empty. We map the 50 most frequent relations to ConceptNet relations following Table~\ref{tab:mapping}, and discard the small number of remaining statements.

\subsection{Implementation}
\label{ssec:implementation}

We implement the proposed selection of commonsense knowledge from Wikidata by using the Knowledge Graph ToolKit (KGTK) \cite{ilievski2020kgtk}. KGTK allows us to carry out the proposed approach in a direct and simple way, despite the challenging size and complexity of Wikidata. The full experiment reported in this paper is coded as three Jupyter Notebooks which run on a laptop in under an hour.\footnote{\url{https://github.com/usc-isi-i2/cskg/tree/master/wikidata}} The starting point is the entire Wikidata split into three Wikidata files in KGTK tabular format (an edge file, a node file, and a qualifiers file), as pre-computed with the \emph{import-wikidata} command.

The concrete steps are as follows. We use a customized Python function to create a subset of the node file that contains only concept nodes, by removing nodes whose labels are either empty or contain a capital letter. 
We use the \emph{ifexists} join operator to filter out edges that do not connect two concepts from the edge file.
The command \emph{remove-columns} trims all columns which are not necessary for the experiment. After this, we run \emph{compact} to remove duplicate edges. At this point, we have a subset of edges that are about concepts (P1). To prepare for the usage filtering and help human readability, we expand the set of columns with the \emph{lift} command to include the labels of the subject, the object, and the relation. We use the aforementioned threshold-based filter to select edges for which both the subject and the object are common concepts. Next, we inspect the remaining edges in terms of their relations. We apply the manual mapping of the top 50 relations (Section~\ref{ssec:approach}) to consolidate the remaining Wikidata graph and make its edge types compatible with the format of CSKG.

These steps produce the subset of Wikidata (\textit{Wikidata-CS}), which satisfies our principles (P1-P3), in the CSKG format. Finally, we use \emph{graph-statistics} to compute metrics over this subset. Wikidata-CS is available for download.\footnote{\url{https://doi.org/10.5281/zenodo.3983029}}

\section{Analysis}

\subsection{General Statistics}
\label{ssec:stats}

Wikidata-CS consists of 71,243 nodes and 106,103 edges. It uses 44 edge types to describe these edges. The mean node degree is 2.98, which is higher than in the \textit{subclass of} subset of Wikidata (2.45)~\cite{ilievski2020consolidating}.  The nodes with the highest PageRank in the resulting graph are: \emph{artificial entity} (\texttt{Q16686448}), \emph{kinship} (\texttt{Q171318}), and \emph{class} (\texttt{Q16889133}), which are more customary 
compared to the top nodes in the unfiltered subclass-of data, all of which describe biochemical concepts~\cite{ilievski2020consolidating}. 

\begin{table}[!t]
	\centering

	\caption{Comparison of the size of Wikidata-CS to commonsense sources in CSKG.}
	\begin{tabular} {c c c c c c c c}
		\toprule
		 & \bf ATOMIC & \bf Concept & \bf Frame & \bf Roget & \bf Visual & \bf Word & \bf Wikidata-CS  \\
		 & & \bf Net & \bf Net & & \bf Genome & \bf Net & \bf (this paper) \\
		 \midrule
		\bf \# nodes & 304,909 & 1,787,373 & 36,582 & 71,804 & 10,830 & 91,294 & \bf 71,243 \\
		\bf \# edges & 732,723 & 3,423,004 & 79,060 & 1,403,461 & 2,218,868 & 111,276 & \bf 106,103\\
		\bottomrule
	\end{tabular}
	\label{tab:sizes}
\end{table}

The five most frequent relations in Wikidata-CS are: \textit{subclass of} (\texttt{P279}), \textit{instance of} (\texttt{P31}), \textit{different from} (\texttt{P1889}), \textit{part of} (\texttt{P361}), and \textit{has part} (\texttt{P527}). The first two account for 68.8\% of all edges, indicating the the commonsense knowledge in Wikidata mostly concerns taxonomic information. 
After mapping the relations to ConceptNet, all commonsense knowledge corresponds to 15 edge types. We perform de-duplication to consolidate edges that were expressed with relations of the same group (e.g., \textit{subclass of} and \textit{instance of}), or in two directions with inverse properties (e.g., \textit{has cause} and \textit{has effect}). The distribution of knowledge across these types is shown in Table~\ref{tab:time} (last column). The final set has 101,771 edges, which is below 0.01\% of the full Wikidata. Next, we compare the content and size of Wikidata-CS to those of other commonsense KGs.

\subsection{Comparison to other graphs in CSKG}
\label{ssec:cskg_graphs}

The integration of Wikidata-CS into CSKG allows one to easily compare its content to other sources, such as ConceptNet. How does the size of the commonsense subset in Wikidata compare to that of the other sources? How much of the commonsense knowledge in Wikidata is already present in these other sources? How much is missing? Conversely: how many edges are defined in ConceptNet or WordNet, but are lacking in Wikidata? 
We provide insight into these questions.

\begin{table}[!t]
	\centering
	\caption{Temporal evolution of the Wikidata commonsense knowledge.}
	\begin{tabular} {c c c c}
		\toprule
		 & \bf 2017-12-27 & \bf 2018-12-10 & \bf 2020-05-04  \\
		 \midrule
\bf /r/IsA            & 31,668 & 45,606 (144\%) & 72,707 (230\%)  \\
\bf /r/PartOf          &  3,390 & 4,416 (130\%) & 7,938 (234\%) \\
\bf /r/HasContext   & 1,968 & 3,189 (162\%) & 6,152 (313\%) \\
\bf /r/DistinctFrom  & 782   & 2,011 (257\%) & 4,934 (631\%)  \\
\bf /r/HasPrerequisite & 413  & 1,965 (476\%) & 4,131 (1,000\%) \\
\bf /r/UsedFor         & 735 & 1,215 (165\%) & 2,469 (336\%) \\
\bf /r/Antonym          & 1,109  & 1,530 (138\%) & 2,184 (197\%) \\
\bf /r/MadeOf    &   415   & 834 (201\%) & 1,426 (344\%)   \\
\bf /r/Synonym    &   478    & 655 (137\%) & 1,070 (224\%)   \\
\bf /r/HasProperty    & 339   & 650 (192\%) & 1,049 (309\%)  \\
\bf /r/Causes   &    150    & 238 (159\%) & 651 (434\%)   \\
\bf /r/DerivedFrom    & 190 & 293 (154\%) & 540 (284\%)  \\
\bf /r/SimilarTo   &  28    & 77 (275\%) & 345 (1,232\%)  \\
\bf /r/CreatedBy    &   51  & 68 (133\%) & 187 (367\%) \\
\bf /r/RelatedTo    &  33  & 40 (121\%) & 42 (127\%)  \\
\hline
		\bf  edges (Wikidata-CS) & 41,769 & 62,787 (150\%) & 101,771 (244\%) \\
		\bf edges (Wikidata) & 405,081,219  & 696,605,955 (172\%) & 1,105,944,515 (273\%) \\ 
		\bottomrule
        \bf nodes (Wikidata-CS) & 32,620 & 47,056 & 71,243 \\
		\bf nodes (Wikidata) & 42,187,222 & 53,004,762 & 84,601,621 \\ 
		\bottomrule
	\end{tabular}
	\label{tab:time}
\end{table}

\begin{table}[!t]
	\centering
	\caption{Overlap between Wikidata and other commonsense knowledge sources.}
	\begin{tabular} {c c c c}
		\toprule
		 \bf Other source & \bf Both & \bf Wikidata-CS only & \bf Other source only  \\
		 \midrule
		\bf ConceptNet & 2,386 & 97,473 (97.6\%) & 3,320,935 (99.9\%) \\
		\bf Roget & 299 & 99,560 (99.7\%) & 1,403,162 (99.9\%) \\
		\bf WordNet & 1,613 & 98,246 (98.4\%) & 419,103 (99.6\%)  \\
		\bottomrule
	\end{tabular}
	\label{tab:stats_overlap}
\end{table}

Table~\ref{tab:sizes} compares the size of Wikidata-CS with the other subgraphs within CSKG. Despite the fact that Wikidata is by far the largest graph, its commonsense subset ranks 6th in terms of edges and 5rd in terms of nodes, being only larger than FrameNet and over 30 times smaller than ConceptNet.\footnote{To be fair, the edge count of the other graphs may include edges with named entities (e.g., through the \texttt{/r/IsA} relation), which were excluded in Wikidata-CS. } 
We also inspect the overlap between the knowledge in Wikidata-CS and in other CSKG sources that share the same relations. Since only the relations are mapped between these sources, whereas the nodes are not, we assume equivalence of two edges with identical subject labels, object labels, and edge types. The results are given in Table \ref{tab:stats_overlap}. We observe that Wikidata-CS shares 2,386 edges with ConceptNet, 1,613 with WordNet, and only 299 with Roget. 
Above all, this investigation shows extremely little overlap between Wikidata-CS and the other three graphs. The observation that commonsense knowledge in Wikidata is almost entirely missing in the other KGs, and vice versa, validates the main pursuit of this paper, and motivates the consolidation of these sources into a single graph.

We note that, with this lexical overlap approach, an edge might be counted multiple times if its nodes have multiple labels. This is why WordNet has over 500k edges in total in Table \ref{tab:stats_overlap}, while having little over 100k in the original data. Future work should investigate more semantic overlap estimation methods.

\subsection{Evolution of the Wikidata commonsense knowledge over time}
\label{ssec:time}

The size of Wikidata has been growing at a tremendous rate. In only 30 months, its number of edges nearly tripled and the number of nodes doubled (Table~\ref{tab:time}). A natural question arises: has the size of its commonsense subset been growing at a similar rate? To investigate this question, we consider three versions of Wikidata, with dates: 2017-12-27, 2018-12-10, and 2020-05-04. For fair comparison, we apply our approach (Section~\ref{ssec:approach}) on the three Wikidata dumps. 

For each of the Wikidata dumps, we present the number of edges per relation in Table~\ref{tab:time}.
Firstly, while the number of edges in Wikidata-CS has multiplied for nearly all relations (except \texttt{RelatedTo}), its growth is slightly slower than the full Wikidata - 244\% vs 273\% between December 2017 and May 2020. A similar trend holds for the December 2018 version. Hence, despite the apparent interest in enriching the commonsense knowledge subset of Wikidata, this has not been a priority so far. 
Secondly, we see larger growth of the relations \textit{SimilarTo}, \textit{HasPrerequisite}, and \textit{DistinctFrom} relative to the others. This shows that certain commonsense aspects (like differentiating potentially confusing concepts) may be more relevant to the Wikidata community and its applications than others. 

\section{Discussion}
\label{ssec:discussion}

The commonsense knowledge in Wikidata could benefit applications like question answering or entity linking. For instance, let's consider the following true/false question from the CycIC dataset:\footnote{\url{https://leaderboard.allenai.org/cycic/submissions/get-started}} \textit{Suppose something is under the table. It is either a toaster or a correction tape dispenser. You can tell that it isn't a kitchen tool. True or False: The thing under the table is a correction tape dispenser}. The key implicit knowledge in this example is the fact that the toaster is often found in the kitchen, while the dispenser is not. Luckily, over 100k such commonsense facts are part of our Wikidata-CS collection, and could help a downstream system to reason over such questions. 

Still, we noted that only a neglegible portion of Wikidata directly describes commonsense knowledge today. Given the considerable community involved in Wikidata and Wikipedia, and the commonsense relations identified in this paper, we propose for the commonsense knowledge in Wikidata to be substantially enriched in the near future. We discuss three actions towards this goal:

\textbf{1. Integration of ready commonsense sources into Wikidata} A number of commonsense sources, like ConceptNet and ATOMIC, contain much complementary knowledge that could be included into Wikidata (cf. Table~\ref{tab:sizes}). Our prior work on consolidating their formats and modeling principles into CSKG enables their seamless integration into Wikidata, when so desired. At present, CSKG contains 5.89 million edges, expressed through 58 relations. The mappings in Table~\ref{tab:mapping} could be used as a starting point, whereas missing relations might need to be added to Wikidata. Data licensing may be a a roadblock here. 

\textbf{2. Generalizing over instance-level knowledge} Much of the commonsense knowledge in Wikidata is indirectly expressed through its instance-level knowledge. While Barack Obama being born in Hawaii is not a commonsense fact, the fact that humans have a birthplace is. Furthermore, all humans have a single birthplace, i.e., it is a functional property. One could think of other generalizations as well, e.g., if many locations belong to countries, it is common sense thinking that any location would belong to a country. Such commonsense information is not directly represented in Wikidata, yet it could be inferred by statistical generalization over instance-level knowledge.

\textbf{3. Missing knowledge types} The Wikidata model defines the notion of qualifiers, which would be ideal to represent much commonsense knowledge. However, 
in many cases, Wikidata does not only lack a commonsense fact, but also the relation or the qualifier that would express it. For instance, while many qualifiers (e.g., minimum and maximum value) express quantities, no qualifier describes typical/expected quantity.\footnote{\url{https://www.wikidata.org/wiki/Wikidata:List\_of\_properties/Wikidata\_qualifier}} This could express that spiders typically have eight legs, while chairs have four. Qualifiers for expressing a purpose or a goal (e.g., one participates in a competition in order to win) are also missing. Besides qualifiers, it might be of use to include relations that are currently missing, like typical properties of concepts (e.g., elephants are heavy), or their symbolism (e.g., red is a symbol of danger). The actual information could be extracted from unstructured sources, like Wikipedia, or reused from previous extractions~\cite{elazar2019large}.









\section{Conclusion}

Wikidata has been growing tremendously in terms of both size and popularity. Consequently, it has been attracting interest from applications that require background knowledge in order to fill in gaps, such as question answering and entity linking. In this paper, we studied the commonsense knowledge coverage of Wikidata and its complementarity to existing commonsense graphs. Starting from three key principles of commonsense, we devised a three-step filtering approach that distinguishes concepts from named entities, favors common concepts, and general-domain knowledge types. Here, we also created mappings between the relations in the commonsense subset of Wikidata (Wikidata-CS) and those in ConceptNet, which allowed us to integrate Wikidata-CS into CSKG, an existing consolidated graph of commonsense knowledge. We analyzed the content of Wikidata-CS and compared it to other existing sources, like ConceptNet and WordNet, noting that while Wikidata contains useful and novel commonsense knowledge that complements other sources, its coverage of commonsense knowledge is currently largely incomplete. We propose three directions to improve this in the future: by inclusion of the knowledge from the Commonsense Knowledge Graph, by generalizing over existing instance-level knowledge in Wikidata, and by inclusion of missing knowledge types that are relevant for representing commonsense knowledge. In addition, subsequent research should evaluate the quality of Wikidata-CS and its relevance for commonsense reasoning, based on user studies and downstream tasks.

\section*{Acknowledgements}

We would like to thank \emph{Daniel Garijo} for the very helpful comments and suggestions on drafts of this paper. We would also like to thank \emph{Craig Milo Rogers} for his help with KGTK.
We are grateful for the reviewer comments.
This material is based upon work sponsored by the DARPA MCS program under Contract No. N660011924033 with the United States Office Of Naval Research and by Air Force Research Laboratory under agreement number FA8750-20-2-10002. 



 \bibliographystyle{splncs04}
 \bibliography{refs}

\end{document}